\pdfoutput=1
\documentclass[11pt]{article}

\usepackage{emnlp2021}

\usepackage{times}
\usepackage{latexsym}

\usepackage[T1]{fontenc}
\usepackage[utf8]{inputenc}

\usepackage{microtype}

\usepackage{amsmath}
\usepackage{multirow}
\usepackage{booktabs}
\usepackage{graphicx}

\title{A Masked Segmental Language Model for Unsupervised Natural Language Segmentation}

\author{C.M. Downey, Fei Xia, Gina-Anne Levow, Shane Steinert-Threlkeld \\
    Department of Linguistics, University of Washington \\
    {\tt \{cmdowney, fxia, levow, shanest\}@uw.edu} \\}
  
\date{}

\begin{document}
\maketitle
\begin{abstract}
 We introduce a Masked Segmental Language Model (MSLM) for unsupervised segmentation. Segmentation remains an important preprocessing step both for languages where ``words'' or other linguistic units (like morphemes) are not clearly delineated by whitespace, as well as for continuous speech data, where there is often no meaningful pause between words. Near-perfect supervised methods have been developed for use in resource-rich languages such as Chinese, but many of the world's languages are both morphologically complex, and have no large dataset of ``gold'' segmentations. To solve this problem, we propose a new type of Segmental Language Model 
 % \citep{sun_unsupervised_2018, kawakami_learning_2019, wang_unsupervised_2021}
 for use in both unsupervised and lightly supervised segmentation tasks. The MSLM is built on a span-masking transformer architecture, harnessing the power of a bidirectional masked context and attention. In a series of experiments, our model outperforms Recurrent SLMs on Chinese in segmentation quality, and performs similarly to the recurrent model on English. We conclude by discussing the challenges posed by segmentation in differing natural language settings.
\end{abstract}

\section{Introduction}
Outside of the orthography of English and languages with similar writing systems, natural language is not usually overtly segmented into meaningful units. Many languages, like Chinese, are written with no spaces in between characters, and Chinese Word Segmentation remains an active field of study (e.g. \citealp{tian_improving_2020}). Running speech is also usually highly fluent with no meaningful pauses existing between ``words'' like in orthography.

Tokenization schemes for powerful modern language models have now largely been passed off to greedy information-theoretic algorithms like Byte-Pair Encoding \citep{sennrich_neural_2016} and the subsequent SentencePiece \citep{kudo_sentencepiece_2018}, which create subword vocabularies of a desired size by iteratively joining commonly co-occuring units. However, these segmentations are usually not sensical to human readers: for instance, \textit{twice} is sometimes modeled as \texttt{tw + ice}, even though a human would know \textit{twice} is not related to \textit{ice}. Given the current performance of models using BPE-type tokenization, the nonsensical nature of these segmentations does not necessarily seem to inhibit the success of neural models.

Nevertheless, BPE does not necessarily help in situations where knowing a sensical segmentation of linguistic-like units is important, such as attempting to model the ways in which children acquire language \citep{goldwater_bayesian_2009}, segmenting free-flowing speech \citep{kamper_unsupervised_2016, rasanen_unsupervised_2020}, creating linguistic tools for morphologically complex languages \citep{moeng_canonical_2021}, or studying the structure of an endangered language  with few or no current speakers \citep{dunbar_zero_2020}.

In this paper, we present a new unsupervised model for robust segmentation. While near-perfect supervised models have been developed for resource-rich languages like Chinese, most of the world's languages do not have large corpora of training data \citep{joshi_state_2020}. Especially for morphologically complex languages, large datasets containing ``gold'' segmentations into units like morphemes are very rare. To solve this problem we propose a type of Segmental Language Model \citep{sun_unsupervised_2018, kawakami_learning_2019}, based on the powerful neural transformer architecture \citep{vaswani_attention_2017}.

Segmental Language Models (SLMs) produce strong language models that can also be used for unsupervised segmentation that correlates with units like words and morphemes \citep{kawakami_learning_2019}. SLMs can be considered a form of character or open-vocabulary language model in which the input is a sequence of characters, and the sequence is modeled as a latent series of segments comprised of one or more characters. 
% The loss term for these models is the marginal probability of all segmentation paths through the sequence. Decoding can then be carried out via efficient dynamic programming.

In departure from previous SLMs, we present a Masked Segmental Language Model (MSLM), built on the transformer's powerful ability to mask out and predict spans of input characters, and embracing a fully bidirectional modeling context with attention. As far as we are aware, we are the first to introduce a non-recurrent architecture for segmental modeling, and conduct comparisons to recurrent baselines across two standard word-segmentation datasets in Chinese and English, with the hope of expanding to more languages and domains (such as speech) in future work.

In Section~\ref{related_work}, we overview baselines in unsupervised segmentation as well as other precursors to the Segmental Language Model. In Section~\ref{model}, we provide a formal characterization of SLMs in general, as well as the architecture and modeling assumptions that make the MSLM distinct. In Section~\ref{experiments}, we present the experiments conducted in this study using both recurrent and masked SLMs, and in Sections~\ref{results}-\ref{analysis} we show that the MSLM outperforms its recurrent counterpart on Chinese segmentation, and performs similarly to the recurrent model on English.
These results demonstrate the power of MSLMs.  Section~\ref{conclusion} lays out directions for future work.
% We consider these results to constitute a proof-of-concept for segmentation with MSLMs (especially for Chinese), and end in Section~\ref{conclusion} by laying out directions for future work with these models.

\section{Related Work}\label{related_work}

\paragraph{Segmentation Techniques and SLM Precursors}
An early application of machine learning to unsupervised segmentation is \citet{elman_finding_1990}, who shows that temporal surprisal peaks of RNNs provide a useful heuristic for inferring word boundaries.  Subsequently, Minimum Description Length (MDL) \citep{rissanen_stochastic_1989} was widely used. The MDL model family underlies well-known segmentation tools such as \textit{Morfessor} \citep{creutz_unsupervised_2002} and other notable works on unsupervised segmentation \citep{de_marcken_linguistic_1996, goldsmith_unsupervised_2001}.

More recently, Bayesian models inspired by child language acquisition have proved some of the most accurate in terms of their ability to model word boundaries. Some of the best examples are Hierarchical Dirichlet Processes \citep{teh_hierarchical_2006}, e.g. those applied to natural language by \citet{goldwater_bayesian_2009}, as well as Nested Pitman-Yor \citep{mochihashi_bayesian_2009, uchiumi_inducing_2015}. However, \citet{kawakami_learning_2019} notes most of these do not adequately account for long-range dependencies in the same capacity as modern neural language models.

Segmental Language Models follow a variety of recurrent models proposed for finding hierarchical structure in sequential data. Influential among these are Connectionist Temporal Classification \citep{graves_connectionist_2006}, Sleep-Wake Networks \citep{wang_sequence_2017}, Segmental RNNs \citep{kong_segmental_2016}, and Hierarchical Multiscale Recurrent Neural Networks \citep{chung_hierarchical_2017}.

In addition, SLMs draw heavily from character and open-vocabulary language models.
% , which can model language units as combinations of subword symbols.
For example, \citet{kawakami_learning_2017} and \citet{mielke_spell_2019} present open-vocabulary language models in which words are represented either as atomic lexical units, or built out of characters. While the hierarchical nature and dual-generation strategy of these models did influence SLMs \citep{kawakami_learning_2019}, both assume that word boundaries are available during training, and use them to form word embeddings from characters on-line.
%\footnote{This in turn is based on \citet{sordoni_hierarchical_2015}, which instead of modeling words as a sequence of characters, models utterance-actions as a sequence of words}.
In contrast, SLMs usually assume no word boundary information is available in training.

\paragraph{Segmental Language Models}
The next section has a more technical description of SLMs; here we give an short overview of other related work. The term Segmental Language Model seems to be jointly due to \citet{sun_unsupervised_2018} and \citet{kawakami_learning_2019}. \citet{sun_unsupervised_2018} demonstrate strong results for Chinese Word Segmentation using an LSTM-based SLM and greedy decoding, competitive with and sometimes exceeding state of the art for the time. This study tunes the model for segmentation quality on a validation set, which we will call a ``lightly supervised'' setting (Section~\ref{supervision_setting}).

\citet{kawakami_learning_2019} use LSTM-based SLMs in a strictly unsupervised setting in which the model is only trained to optimize language-modeling performance on the validation set, and is not tuned on segmentation quality. Here they report that ``vanilla'' SLMs give sub-par segmentations unless combined with one or more regularization techniques, including a character $n$-gram ``lexicon'' and length regularization.

Finally, \citet{wang_unsupervised_2021} very recently introduce a bidirectional SLM based on a Bi-LSTM. They show improved results over the unidirectional SLM of \citet{sun_unsupervised_2018}, test over more supervision settings, and include novel methods for combining decoding decisions over the forward and backward directions. This study is most similar to our own work, though transformer-based SLMs utilize a bidirectional context in a qualitatively different way, and do not require an additional layer to capture the reverse context.

\section{Model}\label{model}
\subsection{Recurrent SLMs}
\begin{figure}[ht]
    \begin{center}
    \includegraphics[width=0.5\textwidth]{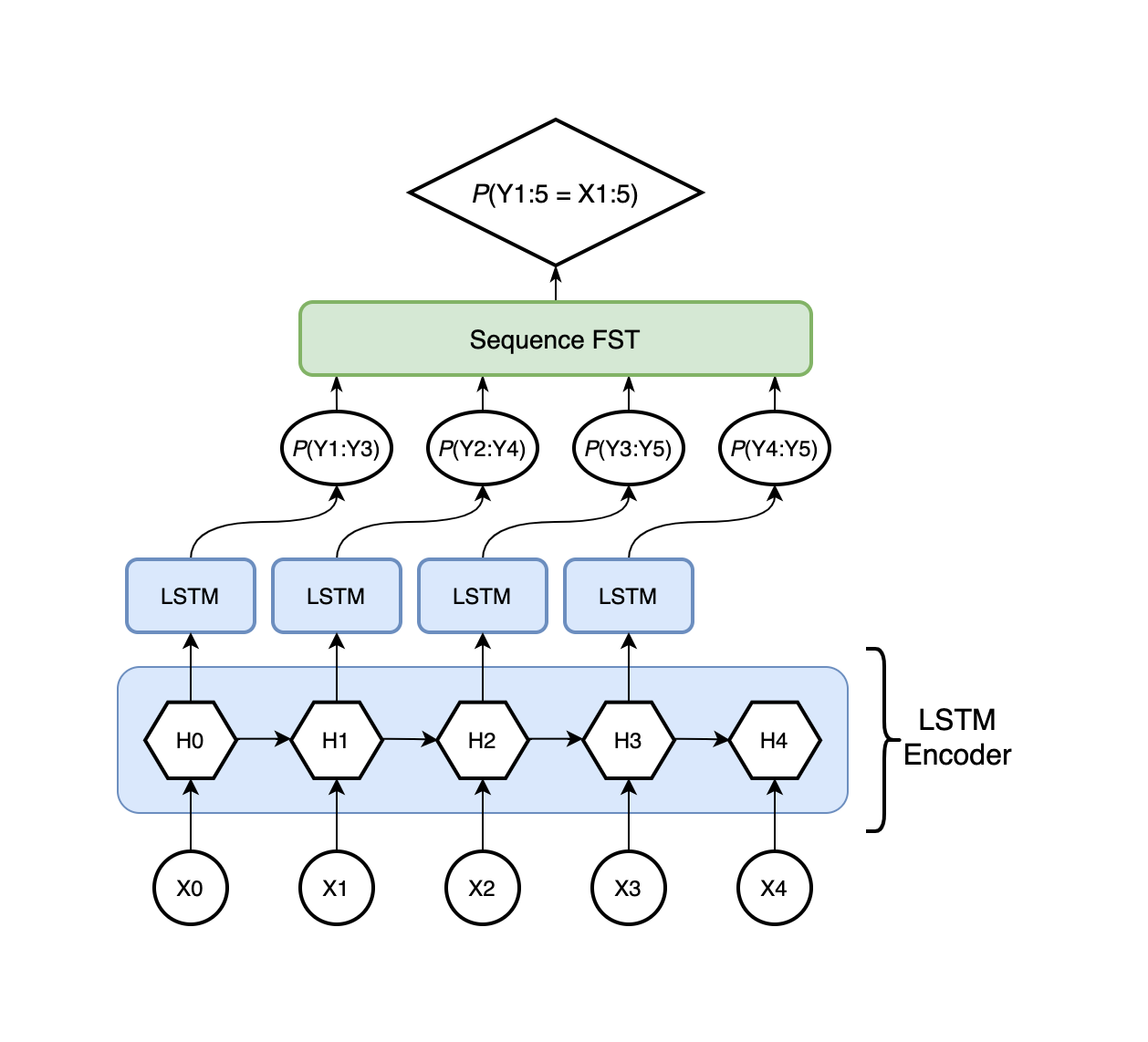}
    \caption{Recurrent Segmental Language Model}
    \label{rslm}
    \end{center}
\end{figure}

% Our model is a type of Segmental Language Model, following the terminology of \citet{sun_unsupervised_2018, kawakami_learning_2019}.
A schematic of the original Recurrent SLM can be found in Figure~\ref{rslm}. Within an SLM, a sequence of symbols or time-steps $\textbf{x}$ can further be modeled as a sequence of segments $\textbf{\underline{y}}$, which are themselves sequences of the input time-steps, such that the concatenation of segments $\pi(\textbf{\underline{y}}) = \textbf{x}$.

SLMs are broken into two levels: a Context Encoder and a Segment Decoder. The Segment Decoder estimates the probability of the $j^{th}$ character in the segment starting at index $i$, $y_j^i$, as: % in the following equation
\[
    p(y_j^i|y_{0:j}^i, x_{0:i}) = \mathit{Decoder}(h_{j-1}^i, y_{j-1}^i)
\]
where the indices for $x_{i:j}$ are $[i,j)$. The Context Encoder encodes information about the input sequence up to index $i$. The hidden encoding $h_i$ is
\[
    h_i = \mathit{Encoder}(h_{i-1}, x_{i})
\]

Finally, the Context Encoder ``feeds'' the Segment Decoder: the initial character of a segment beginning at $i$ is decoded using (transformations of) the encoded context as initial states ($g_h(x)$ and $g_{start}(x)$ are single feed-forward layers):
\begin{align*}
    p(y_0^i|x_{0:i}) &= \mathit{Decoder}(h_{\emptyset}^i, start^i) \\
    h_{\emptyset}^i &= g_{h}(h_{i-1}) \\
    \mathit{start}^i &= g_{\mathit{start}}(h_{i-1})
\end{align*}

For inference, the probability of a segment $\textbf{y}_{i:i+k}$ (starting at index $i$ and of length $k$) is modeled as the log probability of generating $\textbf{y}_{i:i+k}$ with the Segment Decoder given the left context $\pi(\textbf{\underline{y}}_{0:i}) = x_{0:i}$. Note that the probability of a segment is \textbf{not} conditioned on other segments / segmentation choice, but only on the unsegmented input timeseries. Thus, the probability of the segment is
\[
    p(y_0^i|h_{\emptyset}^i,\mathit{start}^i)\displaystyle\prod_{j=1}^{k} p(y_j^i|h_{j-1}^i, y_{j-1}^i)
\]
where $y_{k}^i$ is the end-of-segment symbol.

The probability of a sentence is thus modeled as the marginal probability over all possible segmentations of the input, as in equation (\ref{naive_marginal}) below (where $Z(|\textbf{x}|)$ is the set of all possible segmentations of an input $\textbf{x}$). However, since there are $2^{|\textbf{x}|-1}$ possible segmentations, directly marginalizing is intractable. Instead, dynamic programming over a forward-pass lattice can be used to recursively compute the marginal as in (\ref{dynamic_marginal}) given the base condition that $\alpha_0 = 1$. The maximum-probability segmentation can then be read off of the backpointer-augmented lattice through Viterbi decoding.
\begin{align}
\label{naive_marginal}
    p(\textbf{x}) &= \displaystyle\sum_{\textbf{z} \in Z(|\textbf{x}|)}\displaystyle\prod_i p(\textbf{y}_{i:i+z_i})
    \\
\label{dynamic_marginal}
    p(\textbf{x}_{0:i}) = \alpha_i &= \displaystyle\sum_{k=1}^L p(\textbf{y}_{i-k:i}|\textbf{x}_{0:i-k}) \alpha_{i-k}
\end{align}

\subsection{New Model: Masked SLM}\label{sec_mslm}
We present a Masked Segmental Language Model, which leverages a non-directional transformer as the Context Encoder.  This reflects recent advances in bidirectional \citep{schuster_bidirectional_1997, graves_framewise_2005, peters_deep_2018} and adirectional architectures in language modeling \citep{devlin_bert_2019}. Such modeling contexts are also psychologically plausible: \citet{luce_computational_1986} shows that in acoustic perception, most words need some following context to be recognizable.

A key difference between our model and standard, transformer-based Masked Language Models like BERT is that the latter predict single tokens based on the rest, while for SLMs we are interested in predicting a \emph{segment} of tokens based on all other tokens \textit{outside the segment}. For instance, to predict the three-character segment starting at $x_t$, the distribution to be modeled is
    $ p(\textbf{x}_{t:t+3}|\textbf{x}_{<t}, \textbf{x}_{\geq t+3})$.

There are some recent pre-training techniques for transformers, such as MASS \citep{song_mass_2019} and BART \citep{lewis_bart_2020}, that mask out spans to be predicted. One key difference between our model and these approaches is that while the pre-training data for large transformer models is usually large enough that only about 15\% of training tokens are masked, we will always want to estimate the generation probability for \textit{every} possible segment of $\textbf{x}$. Since the usual method for masking requires replacing the masked token(s) with a special symbol, only one span can be predicted with each forward pass. However, in each sequence there are $O(|\textbf{x}|)$ possible segments, so replacing each one with a mask token and recovering it would require as many forward passes.

These design considerations motivate our \textbf{Segmental Transformer Encoder}, and the \textbf{Segmental Attention Mask} around which it is based. For each forward pass of this encoder, an encoding is generated for every possible starting position in $\textbf{x}$ for a segment of up to length $k$. The encoding at timestep $t-1$ corresponds to every possible segment whose first timestep is at index $t$. Thus with maximum segment length of $k$ and total sequence length $n$, the encoding at each index $t-1$ will approximate
\[
    p(\textbf{x}_{t:t+1}, \textbf{x}_{t:t+2}, ...\textbf{x}_{t:t+k} | \textbf{x}_{<t}, \textbf{x}_{\geq t+k})
\]

\begin{figure}[ht]
    \begin{center}
    \includegraphics[width=0.25\textwidth]{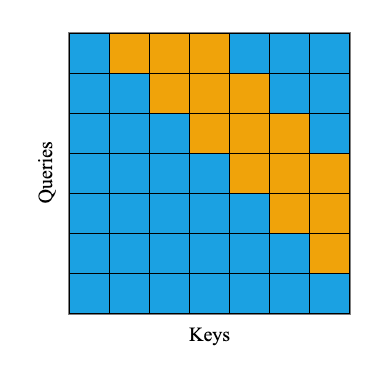}
    \caption{Segmental Attention Mask with segment-length ($k$) of 3. Blue squares are equal to $0$, orange squares are equal to $-\infty$. This mask blocks the position encoding the segment in the Queries from attending to segment-internal positions in the Keys.}
    \label{mask}
    \end{center}
\end{figure}

This encoder leverages an attention mask designed to condition predictions only on indices that are not part of the segment to be predicted. An example of this mask with $k=3$ is shown in Figure \ref{mask}. For max segment length $k$, the mask is given by:
\[
    \alpha_{i,j} = 
    \begin{cases} 
        -\infty & \text{if } 0 < j-i \leq k \\
        0 & \text{else}
    \end{cases}
\]

This solution is similar to that of \citet{shin_fast_2020}, developed independently and concurrently with the present work, which uses a custom attention mask to ``autoencode'' each position in the sequence without the need to input a special mask token. One key difference is that their masking scheme is used to predict single tokens, rather than spans. In addition, their mask runs directly along the diagonal of the attention matrix, rather than being offset. This means that to preserve self-masking in the first layer, the Queries are the ``pure'' positional embeddings, rather than a transformation of the input sequence.

To prevent information leaking ``from under the mask'', our encoder uses a different configuration in its first layer than in subsequent layers. In the first layer, Queries, Keys, and Values are all learned from the original input encodings. In subsequent layers, the Queries come from the hidden encodings output by the previous layer, while Keys and Values are learned directly from the original encodings. If Queries and Keys or Queries and Values both come from the previous layer, information can leak from positions that are supposed to be masked for each respective query position. \citet{shin_fast_2020} come to a similar solution to preserve their auto-encoder masking.

\begin{figure}[ht]
    \begin{center}
    \includegraphics[width=0.5\textwidth]{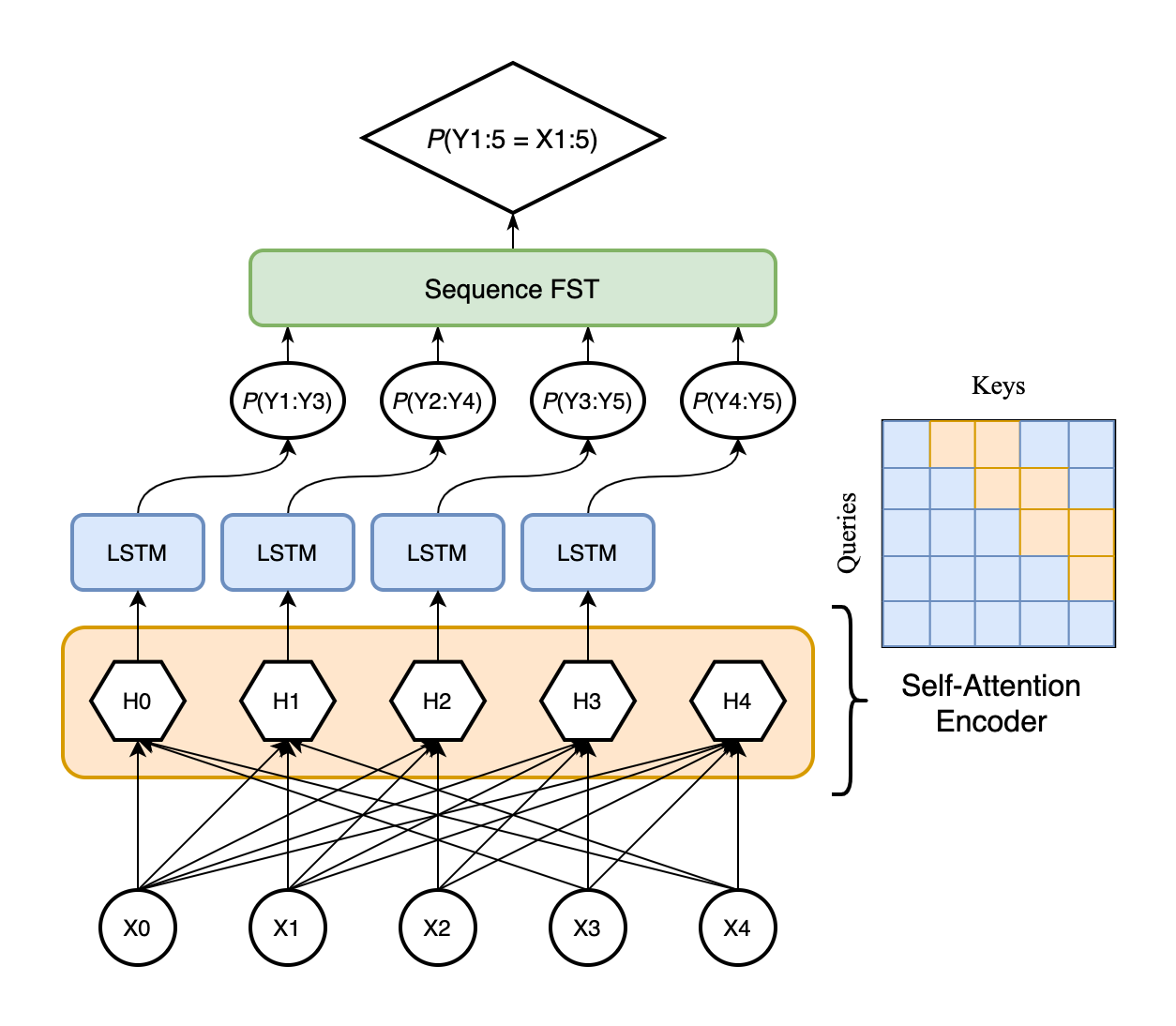}
    \caption{Masked Segmental Language Model, $k=2$.}
    \label{mslm}
    \end{center}
\end{figure}

The encodings learned by the segmental encoder can then be input to an SLM decoder in exactly the same way as previous models (Figure \ref{mslm}).

% Switching to a transformer-based SLM makes the language modeling context naturally bidirectional (or rather adirectional). However, a transformer encoder does not necessarily have to be bidirectional, taking for example, the directional or ``causal'' mask used for sequence decoding with transformers \citep{vaswani_attention_2017}. In addition, an RNN-based encoder can be bidirectional (although the latent representations of each direction are separate and need to be combined; see \citealp{wang_unsupervised_2021} for a BiLSTM-based SLM).

\begin{figure}[ht]
    \begin{center}
    \includegraphics[width=0.5\textwidth]{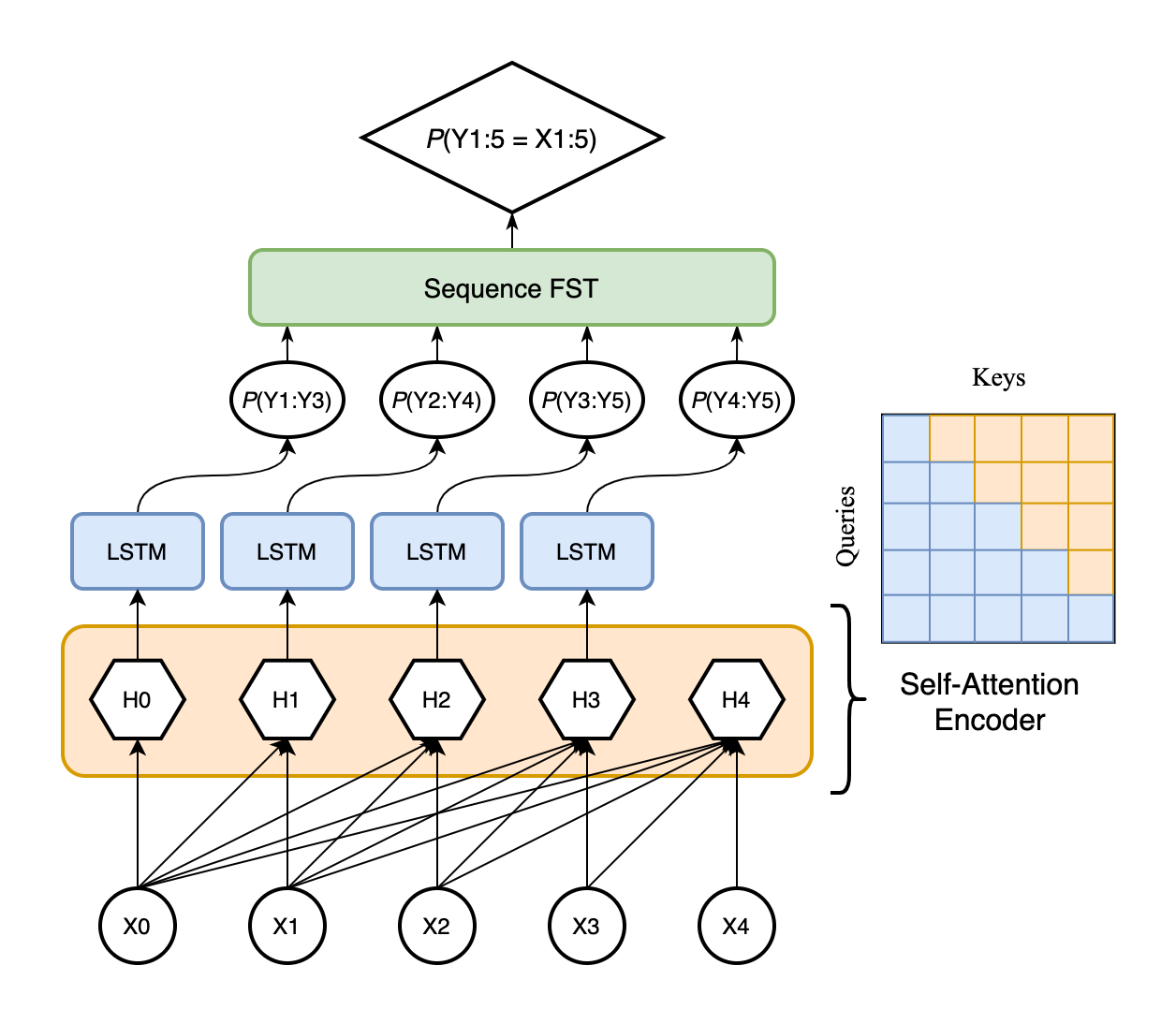}
    \caption{Directional MSLM}
    \label{dmslm}
    \end{center}
\end{figure}

To tease apart the role of attention in a transformer and the encoder's adirectional context modeling assumption, we additionally define a Directional Masked Segmental Language Model, which uses a directional mask instead of the span masking type. Using the directional mask, the encoder is still completely attention-based, but the language modeling context is strictly ``directional'', in that positions are only allowed to attend over a monotonic ``leftward'' context (Figure \ref{dmslm}).

Finally, to add positional information to the  encoder, we use static sinusoidal encodings \citep{vaswani_attention_2017} and additionally employ a linear mapping $f$ to the concatenation of the original and positional embeddings to learn the ratio at which to add the two together.
\begin{align*}
    g = 1.0 + \mathit{ReLU}(f([\mathit{embedding, position}]))
    \\ \mathit{embedding} \xleftarrow{} g * \mathit{embedding} + \mathit{position}
\end{align*}

\section{Experiments}\label{experiments}
Our experiments assess SLMs across three dimensions: (1) network architecture and language modeling assumptions, (2) evaluation metrics, specifically segmentation quality and language-modeling performance, and (3) supervision setting (if and where gold segmentation data is available).

\subsection{Architecture and Modeling}
To analyze the importance of the self-attention architecture versus the bidirectional conditioning context, we test SLMs with three different encoders: the standard R(ecurrent)SLM based on an LSTM, the M(asked)SLM introduced in \ref{sec_mslm} with a segmental or ``cloze'' mask, and a D(irectional)MSLM, with a ``causal'' or directional mask. The RSLM is thus (+recurrent context, +directional), the DMSLM is (-recurrent context, +directional), and the MSLM is (-recurrent context, -directional).

For all models, we use an LSTM for the segment decoder, as a control and because the decoded sequences are relatively short and may not benefit as much from an attention model. See also \citet{chen_best_2018} for hybrid models with transformer encoders and recurrent decoders.

\subsection{Evaluation Metrics}
% There are many segmentation models that are not strong language models, e.g. the Bayesian models mentioned in Section~\ref{related_work}. 
Part of the motivation for SLMs is to create strong language models that can also be used for segmentation \citep{kawakami_learning_2019}. Because of this, we report both segmentation quality and language modeling performance.

For segmentation quality, we get the word-F1 score for each corpus using the script from the SIGHAN Bakeoff \citep{emerson_second_2005}. Following \citet{kawakami_learning_2019}, we report this measure over the entire corpus. For language modeling performance, we report the average Bits Per Character (bpc) loss over the test set.

\subsection{Supervision Setting}\label{supervision_setting}
Because previous studies have used SLMs both in ``lightly supervised'' settings \citep{sun_unsupervised_2018} and totally unsupervised ones \citep{kawakami_learning_2019}, and because we expect SLMs to be deployed in either use case, we test both. For all model types, we conduct a hyperparameter sweep and select both the configuration that maximizes the validation segmentation quality (light supervision) and the one that minimizes the validation bpc (unsupervised).

\subsection{Datasets}
We evaluate our SLMs on two datasets used in \citet{kawakami_learning_2019}. For each, we use the same training, validation, and test split. The sets were chosen to represent two relatively different writing systems: Chinese (PKU) and English (PTB). Statistics for each are in Table~\ref{dataset_stats}. One striking difference between the two writing systems can be seen in the character vocabulary size: phonemic-type writing systems like English have a much smaller vocabulary of tokens, with words being built out of longer sequences of characters that are not meaningful on their own.
%We speculate on the effects each of these systems has on modeling performance in Section~\ref{analysis}.
\begin{table}[ht]
    \centering
    \begin{tabular}{lcc}
        \toprule
        Corpus & PKU & PTB \\
        \midrule
        Tokens/Characters & 1.93M & 4.60M \\
        Words & 1.21M & 1.04M \\
        Lines & 20.78k & 49.20k \\
        Avg. Characters per Word & 1.59 & 4.44 \\
        Character Vocabulary Size & 4508 & 46 \\
        \bottomrule
    \end{tabular}
    \caption{Statistics for the datasets}
    \label{dataset_stats}
\end{table}

\paragraph{Peking University Corpus (PKU)} PKU has been used as a Chinese Word Segmentation benchmark since the International Chinese Word Segmentation Bakeoff \citep{emerson_second_2005}. One minor change we make to this dataset is to tokenize English, number, and punctuation tokens using the module from \citet{sun_unsupervised_2018}, to make our results more comparable to theirs. Unlike them, we do not pre-split sequences on punctuation.

\paragraph{Penn Treebank (PTB)}
For English, we use the version of the Penn Treebank corpus from \cite{kawakami_learning_2019, mikolov_recurrent_2010}.

\subsection{Parameters and Trials}
For all models, we tune among six learning rates on a single random seed. After the parameter sweep, the configuration that maximizes validation segmentation quality and the one that minimizes validation bpc are run over an additional four random seeds. All models are trained using Adam \citep{kingma_adam_2015} for 8192 steps.

All models have one encoder layer and one decoder layer, as well as an embedding and hidden size of 256. The transformer-based encoder has a number of trainable parameters less than or equal to the number in the LSTM-based encoder.\footnote{592,381 trainable parameters in the former, 592,640 in the latter}

One important parameter for SLMs is the maximum segment length $k$. \citet{sun_unsupervised_2018} tune this as a hyperparameter, with different values for $k$ fitting different Chinese segmentation standards more or less well. In practice, this parameter can be chosen empirically to be an upper bound on the maximum segment length one expects to find, so as to not rule out long segments. We follow \citet{kawakami_learning_2019} in choosing $k=5$ for Chinese and $k=10$ for English.
For a more complete characterization of our training procedure, see Appendix~\ref{app:training_deets}.\footnote{The code used to build SLMs and conduct these experiments can be found at \url{https://github.com/cmdowney88/SegmentalLMs}}

\section{Results}\label{results}
\subsection{Chinese}
\begin{table*}[ht]
    \centering
    \begin{tabular}{clcccc}
        \toprule
        \multirow{2}{*}{Dataset} & \multirow{2}{*}{Model} & \multicolumn{2}{c}{Tuned on Gold} & \multicolumn{2}{c}{Unsupervised}\\
        \cmidrule{3-4} \cmidrule{5-6}
        {} & {} & F1 Mean / Median & BPC & F1 Mean / Median & BPC \\
        \midrule
        \multirow{3}{*}{PKU} & RSLM & 61.2 $\pm$ 3.6 / 60.2 & \textbf{5.67 $\pm$ 0.01} & 59.4 $\pm$ 1.9 / 58.7 & 5.63 $\pm$ 0.01 \\
        {} & DMSLM & 72.2 $\pm$ 2.0 / 72.7 & 6.08 $\pm$ 0.31 & 62.9 $\pm$ 2.6 / 63.4 & 5.67 $\pm$ 0.03 \\
        {} & MSLM & \textbf{72.3 $\pm$ 0.7 / 72.6} & 5.85 $\pm$ 0.12 & \textbf{62.9 $\pm$ 2.8 / 64.1} & \textbf{5.56 $\pm$ 0.01} \\
        \midrule
        \multirow{3}{*}{PTB} & RSLM & \textbf{77.4 $\pm$ 0.7 / 77.6} & \textbf{2.10 $\pm$ 0.04} & \textbf{75.7 $\pm$ 2.6 / 76.2} & \textbf{1.96 $\pm$ 0.00} \\
        {} & DMSLM & 70.6 $\pm$ 6.4 / 73.3 & 2.36 $\pm$ 0.07 & 67.9 $\pm$ 10.6 / 73.8 & 2.27 $\pm$ 0.04 \\
        {} & MSLM & 71.1 $\pm$ 5.6 / 73.6 & 2.39 $\pm$ 0.06 & 69.3 $\pm$ 5.6 / 71.5 & 2.27 $\pm$ 0.01 \\
        \bottomrule
    \end{tabular}
    \caption{Results on the Peking University Corpus and English Penn Treebank (over 5 random seeds)}
    \label{results_table}
\end{table*}

For PKU (Table \ref{results_table}), Masked SLMs yield better segmentation quality in both the lightly-supervised and unsupervised settings, though the advantage in the former setting is much larger (+12.4 median F1). The Directional MSLM produces similar quality segmentations to the MSLM, but it has worse language modeling performance in both settings (+0.23 bpc for lightly supervised and +0.11 bpc for unsupervised); the RSLM produced the second-best bpc in the unsupervised setting.

The RSLM gives the best bpc in the lightly-supervised setting. However for this setting, the strict division of the models that maximize segmentation quality and those that minimize bpc can be misleading. In between these two configurations, many have both good segmentation quality and low bpc, and if the practitioner has gold validation data, they will be able to pick a configuration with the desired tradeoff.

In addition, there is some evidence that ``undershooting'' the objective in the unsupervised case with a slightly lower learning rate may lead to more stable segmentation quality. The unsupervised MSLM in the table was trained at rate 2e-3, and achieved 5.625 bpc (validation). An MSLM trained at rate 1e-3 achieved only a slightly worse bpc (5.631) and resulted in better and more stable segmentation quality (69.4 $\pm$ 2.0 / 70.4).

\subsection{English}

Results for English (PTB) can also be found in Table~\ref{results_table}. By median, results remain comparable between the recurrent and transformer-based models, but the RSLM yields better segmentation performance in both settings (+4.0 and +4.7 F1). However, both types of MSLM are slightly more susceptible to random seed variation, causing those means to be skewed slightly lower. The DMSLM seems more susceptible than the MSLM to outlier performance based on random seeds, as evidenced by its large standard deviation. Finally, the RSLM gives considerably better bpc performance in both settings (-0.29 and -0.31 bpc).

\section{Analysis and Discussion}\label{analysis}
\subsection{Error Analysis}
We conduct an error analysis for our models based on the total Precision and Recall scores for each (using the character-wise binary classification task, i.e. word-boundary vs no word-boundary).

As can be seen in Table \ref{analysis_table}, all model types trained on Chinese have a Precision that approaches 100\%, meaning almost all boundaries that are inserted are true boundaries. On first glance the main difference in the unsupervised case seems to be the RSLM's relatively higher Recall. However, the higher Precision of both MSLM types seems to be more important for the overall segmentation performance.\footnote{This table also shows that though character-wise segmentation quality (i.e. classifying whether a certain character has a boundary after it) is a useful heuristic, it does not always scale in a straightforward manner to word-wise F1 like is traditionally used (e.g. by the SIGHAN script).} In the lightly-supervised case, the MSLM variants learn to trade off a small amount of Precision for a large gain in Recall, allowing them to capture more of the true word boundaries in the data. Given different corpora have different standards for the coarseness of Chinese segmentation, this reinforces the need for studies on a wider selection of datasets.

Because the English results (also in Table~\ref{analysis_table}) are similar between supervision settings, we only show the unsupervised variants. Here, the RSLM shows a definitive advantage in Recall, leading to overall better performance. The transformer-based models show equal or higher Precision, but tend to under-segment, i.e. produce longer words. Example model segmentations for PTB can be found in Table \ref{ptb_examples}. Some intuitions from our error analysis can be seen here: the moderate Precision of these models yields some false splits like \texttt{be + fore} and \texttt{quest + ion}, but all models also seem to pick up some valid morphological splits not present in the gold standard (e.g. \texttt{+able} in \textit{questionable}). Predictably, rare words with uncommon structure remain difficult to segment (e.g. \textit{asbestos}).

\begin{table*}[ht]
    \centering
    \begin{tabular}{clccc}
        \toprule
        Dataset & Model & Avg. Word Length & Precision & Recall \\
        \midrule
        \multirow{7}{*}{PKU} & Gold & 1.59 & - & - \\
        {} & RSLM (\textit{unsup.}) & 1.93 $\pm$ 0.02 & 98.2 $\pm$ 0.1 & 80.8 $\pm$ 0.6 \\
        {} & DMSLM (\textit{unsup.}) & 1.99 $\pm$ 0.04 & 98.6 $\pm$ 0.1 & 78.5 $\pm$ 1.8 \\
        {} & MSLM (\textit{unsup.}) & 2.00 $\pm$ 0.05 & 98.5 $\pm$ 0.1 & 78.1 $\pm$ 1.9 \\
        {} & RSLM (\textit{sup.}) & 1.92 $\pm$ 0.02 & 98.2 $\pm$ 0.1 & 81.3 $\pm$ 0.7 \\
        {} & DMSLM (\textit{sup.}) & 1.83 $\pm$ 0.04 & 97.5 $\pm$ 0.5 & 84.6 $\pm$ 1.5 \\
        {} & MSLM (\textit{sup.}) & 1.83 $\pm$ 0.01 & 97.6 $\pm$ 0.1 & 84.5 $\pm$ 0.4 \\
        \midrule
        \multirow{4}{*}{PTB} & Gold & 4.44 & - & - \\
        {} & RSLM (\textit{unsup.}) & 4.02 $\pm$ 0.08 & 86.1 $\pm$ 1.9 & 95.5 $\pm$ 0.1 \\
        {} & DMSLM (\textit{unsup.}) & 4.27 $\pm$ 0.17 & 85.4 $\pm$ 5.4 & 88.9 $\pm$ 4.6 \\
        {} & MSLM (\textit{unsup.}) & 4.29 $\pm$ 0.12 & 86.2 $\pm$ 1.5 & 89.5 $\pm$ 3.5 \\
        \bottomrule
    \end{tabular}
    \caption{Error analysis statistics (over 5 random seeds)}
    \label{analysis_table}
\end{table*}

\begin{table*}[ht]
    \centering
    \begin{tabular}{cl}
    \toprule
     & Examples \\
    \midrule
    Gold &  \small{we 're talking about years ago before anyone heard of asbestos having any questionable...} \\
    RSLM Median & \small{\textbf{we're} talking about years ago \textbf{be fore} \textbf{any one} heard of \textbf{as best os} having any \textbf{question able}} \\
    DMSLM Median & \small{\textbf{we're} talking about years ago \textbf{be fore} \textbf{any one} heard of \textbf{as bestos} having any \textbf{quest ion able}} \\
    MSLM Median & \small{\textbf{we're} talking about years ago \textbf{be fore} \textbf{any one} heard of \textbf{as bestos} having any \textbf{quest ion able}} \\
    \bottomrule
    \end{tabular}
    \caption{Example model segmentations from the Penn Treebank}
    \label{ptb_examples}
\end{table*}

\subsection{Discussion}
For Chinese, the transformer-based SLM exceeds the recurrent baseline for segmentation quality, by a moderate amount for the unsupervised setting, and by a large amount when tuned on gold validation segmentations. The MSLM also gives stronger language modeling. Given the large vocabulary size for Chinese, it is intuitive that the powerful transformer architecture may make a difference in this difficult language-modeling task. Further, though the DMSLM achieves similar segmentation quality, the bidirectional context of the MSLM does seem to be the source of the best bpc modeling performance.

In English, on the other hand, recurrent SLMs seem to retain a slight edge. By median, segmentation quality remains fairly similar between the three model types, but the RSLM holds a major language-modeling advantage in our experiments. Our main hypothesis for the disparity in modeling performance between Chinese and English comes down to the nature of the orthography for each. As noted before, Chinese has a much larger character vocabulary. This is because in Chinese, almost every character is a morpheme itself (i.e. it has some meaning). English, on the other hand, has a roughly phonemic writing system, e.g. the letter \textit{c} has no inherent meaning outside of a context like \textit{cat}.

Intuitively, it is easy to see why this might pose a limitation on transformers. Without additive or learned positional encodings, they are essentially adirectional. In English, \textit{cat} is a completely different context than \textit{act}, but this might be difficult to model for an attention model without robust positional information. To try to counteract this, we added dynamic scaling to our static positional encodings, but without deeper networks or more robust positional information, the discrepancy in character-based modeling for phonemic systems may remain.

\section{Conclusion}\label{conclusion}
This study constitutes strong proof-of-concept for the use of Masked Segmental Language Models in unsupervised segmentation and/or character modeling, with several open avenues to extend their utility. To close, we lay out directions for future work with MSLMs, including extension to more languages and domains, applications to deep modeling, and regularization techniques.

The most obvious next step is using MSLMs to model more segmentation datasets. As mentioned previously, the criteria for what defines a ``word'' in Chinese are not agreed upon, and so more experiments are warranted using corpora with different standards. Prime candidates include the Chinese Penn Treebank \citep{xue_penn_2005}, as well as those included in the SIGHAN segmentation bakeoff: Microsoft Research, City University of Hong Kong, and Academia Sinicia \citep{emerson_second_2005}.

The Chinese and English sets examined here are also relatively formal orthographic datasets. An eventual use of SLMs may be in speech segmentation, but a smaller step in that direction could be using phonemic transcript datasets like the Brent Corpus, also used in \citet{kawakami_learning_2019}, which consists of phonemic transcripts of child-directed English speech \citep{brent_efficient_1999}. SLMs could also be applied to the orthographies of more typologically diverse languages, such as ones with complicated systems of morphology (e.g. Swahili, Turkish, Hungarian, Finnish).

Further, though we have only experimented with shallow models here, one of the main advantages of transformers is their ability to scale to deep models due to their short derivational chains. That being so, extending segmental models to ``deep'' settings may be more feasible using MSLMs than RSLMs.

Lastly, \citet{kawakami_learning_2019} propose regularization techniques for SLMs due to low segmentation quality from their ``vanilla'' models. They report good findings using a character $n$-gram ``lexicon'' in conjunction with expected segment length regularization based on \citet{eisner_parameter_2002} and \citet{liang_online_2009}. Both techniques are implemented in our codebase, and we have tested them in pilot settings. Oddly, neither has given us any gain in performance over our ``vanilla'' models tested here. A more exhaustive hyperparameter search with these methods for both kinds of models may produce a future benefits as well.
% Given these techniques introduce several new hyperparameters, it is possible that we were simply unable to find ideal settings for these parameters using our limited ability to sweep configurations. Indeed, one important hyperparameter from \citet{kawakami_learning_2019} is given to three digits of precision. 

In conclusion, the present study shows strong potential for the use of MSLMs. They show particular promise for writing systems with a large inventory of semantic characters (e.g. Chinese), and we believe that they could be stable competitors of recurrent models in phonemic-type writing systems if the relative weakness of positional encodings in transformers can be mitigated.

\bibliography{MSLM}
\bibliographystyle{acl_natbib}

\appendix
\clearpage
\section{Training Details}\label{app:training_deets}
\subsection{Data}
The datasets used here are sourced from \citet{kawakami_learning_2019}, and can be downloaded at \url{https://s3.eu-west-2.amazonaws.com/k-kawakami/seg.zip}. Our PKU data is tokenized slightly differently, and all data used in our experiments can be found in our project repository (url redacted).

\subsection{Architecture}
A dropout rate of 0.1 is applied leading into both the encoder and the decoder. Transformers use 4 attention heads and a feedforward size of 509 (chosen to come out less than or equal to the number of parameters in the standard LSTM). This also includes a 512-parameter linear mapping to learn the combination proportion of the word and sinusoidal positional embeddings. The dropout within transformer layers is 0.15.

\subsection{Initialization}
Character embeddings are initialized using CBOW \citep{mikolov_efficient_2013} on the given training set for 32 epochs, with a window size of 5 for Chinese and 10 for English. Special tokens like \texttt{<eoseg>} that do not appear in the training corpus are randomly initialized. These pre-trained embeddings are not frozen during training.

\subsection{Training}
For PKU, the learning rates swept are \{6e-4, 7e-4, 8e-4, 9e-4, 1e-3, 2e-3\}, and for PTB we use \{6e-4, 8e-4, 1e-3, 3e-3, 5e-3, 7e-3\}. For Chinese, we found a linear warmup for 1024 steps was useful, followed by a linear decay. For English, we apply simple linear decay from the beginning. Checkpoints are taken every 128 steps. A gradient norm clip threshold of 1.0 is used. Mini-batches are sized by number of characters rather than number of sequences, with a size of 8192 (though this is not always exact since we do not split up sequences). The five random seeds used are \{2, 3, 5, 8, 13\}.

Each model is trained on an Nvidia Tesla M10 GPU with 8GB memory, with the average per-batch runtime of each model type listed in Table~\ref{tab:runtime}.

\begin{table}[ht]
    \centering
    \begin{tabular}{ccc}
        \toprule
        \multirow{2}{*}{Model} & \multicolumn{2}{c}{s / step} \\
        \cmidrule{2-3}
        {} & PKU & PTB \\
        \midrule
        RSLM & 2.942 & 2.177 \\
        DMSLM & 2.987 & 2.190 \\
        MSLM & 2.988 & 2.200 \\
        \bottomrule
    \end{tabular}
    \caption{Average runtime per batch in seconds}
    \label{tab:runtime}
\end{table}

\subsection{Optimal Hyperparameters}
The optimal learning rate for each model type, dataset, and supervision setting are listed in the Table~\ref{tab:optimum_parameters}. Parameters are listed by the validation objective they optimize: segmentation MCC or language-modeling BPC.

\begin{table}[ht]
    \centering
    \begin{tabular}{cccc}
        \toprule
        Dataset & Model & by MCC & by BPC \\
        \midrule
        \multirow{3}{*}{PKU} & RSLM & 6e-4 & 9e-4 \\
        {} & DMSLM & 6e-4 & 2e-3 \\
        {} & MSLM & 6e-4 & 2e-3 \\
        \midrule
        \multirow{3}{*}{PTB} & RSLM & 7e-3 & 3e-3 \\
        {} & DMSLM & 1e-3 & 8e-4 \\
        {} & MSLM & 1e-3 & 6e-4 \\
        \bottomrule
    \end{tabular}
    \caption{Optimum learning rates}
    \label{tab:optimum_parameters}
\end{table}

\end{document}